\documentclass[11pt]{article}

\usepackage[margin=1in]{geometry}
\usepackage{amsmath}
\usepackage{amssymb}
\usepackage{booktabs}
\usepackage{hyperref}
\usepackage{xcolor}
\usepackage{graphicx}
\usepackage{enumitem}
\usepackage{natbib}
\usepackage{microtype}
\usepackage[utf8]{inputenc}
\usepackage{newunicodechar}
\newunicodechar{≥}{\ensuremath{\geq}}
\usepackage{multirow}
\usepackage[T1]{fontenc}
\usepackage[utf8]{inputenc}
\usepackage{lineno}
\usepackage{float}
\usepackage{authblk}

\usepackage{listings}
\hypersetup{
  colorlinks=true,
  linkcolor=blue!60!black,
  citecolor=blue!60!black,
  urlcolor=blue!60!black,
}

\title{\textbf{Fine-tuning LLMs for Passive Depression Severity Estimation from AI Mental Health Dialogue}}

\author[1]{Olivier Tieleman}
\author[1]{Samuel J. Bell}
\author[1]{Ziyi Zhu}
\author[1]{Ting Su}
\author[1,*]{Elizabeth Kroll}
\author[1]{Thomas D. Hull}
\author[1]{Caitlin A. Stamatis}
\affil[1]{Slingshot AI, New York, NY, USA}
\affil[*]{Corresponding author: Elizabeth Kroll (\href{mailto:[elizabeth@slingshotai.com]}{[elizabeth@slingshotai.com]})}

\date{\today}

\begin{document}

\maketitle

% ─── Abstract ────────────────────────────────────────────────────────────────
%TC:ignore
\begin{abstract}
Depression is a leading cause of disability worldwide, yet validated questionnaires such as the Patient Health Questionnaire-9 (PHQ-9) often go uncompleted, yielding sparse, biased data. Users of AI mental health applications often describe their mood, sleep and functioning unprompted, producing a continuous record of symptoms. Here, we predict PHQ-9 scores from these conversations alone. As in many clinical language tasks, labels are scarce and imbalanced. We address both by fine-tuning Qwen 3.5 27B with a regression head and augmenting 3,111 ground-truth labels with pseudo-labels from a frontier reasoning model and iteratively trained intermediate models, expanding to 5,524 users with more balanced scores. On 422 held-out users, the model achieved MAE $= 3.98$, Pearson $r = 0.67$, and AUC $= 0.88$ at the clinical threshold (PHQ-9 ≥ 10); AUC was $> 0.80$ across severity thresholds from PHQ-9~$\geq 3$ to PHQ-9~$\geq 24$, demonstrating that the model captures depression severity across the full clinical spectrum. These results support passive depression monitoring without self-report.
\end{abstract}
%TC:endignore
% ─── 1. Introduction ─────────────────────────────────────────────────────────
\section{Introduction}

Depression is the leading cause of disability worldwide \citep{friedrich2017}, affecting more than 280 million people \citep{who2023}, and earlier detection of worsening symptoms is critical for timely intervention. The Patient Health Questionnaire-9 (PHQ-9) is a standardized self-report measure widely used to screen for depression and track its severity over time. Its nine items correspond to the diagnostic criteria for major depressive disorder, including anhedonia, fatigue, feelings of worthlessness, psychomotor changes and suicidal ideation, and yield a total score from 0 to 27. However, the PHQ-9’s value for monitoring depends on repeated administration, which faces two obstacles: declining compliance and non-random attrition. In intensive longitudinal designs, longer and more frequent questionnaires increase perceived burden and careless responding, and compliance declines over time \citep{eisele2022}. The same problem plagues remote digital health research; across eight studies with more than 100,000 participants, median retention was only 5.5 days \citep{pratap2020}. Notably, this attrition is not random with respect to outcome. Participants who were more depressed at baseline were more likely to drop out early, and those whose depression did not improve stopped completing the PHQ-9 sooner than those who improved, such that observed data overstate symptom improvement \citep{hamitouche2025}. Estimating PHQ-9 scores from conversations users already have with AI mental health applications could address both obstacles. It adds no burden and produces an estimate whenever a user engages, not only when they choose to complete a questionnaire.

\subsection{PHQ-9 and Depression Prediction from Text}
Prior work in natural language processing (NLP) has estimated depressive symptom severity from two broad classes of text: language \emph{elicited} within a structured clinical assessment, such as clinical interview transcripts, and \emph{naturalistic} language generated in everyday contexts, such as social media posts or text messages. This distinction has direct implications for deployment. Language elicited during an assessment is typically accompanied by a clinician rating or questionnaire score, such that an automated estimate largely duplicates information already available. Naturalistic language, by contrast, is produced in the absence of formal assessment, which is precisely the setting in which an automated estimate offers the greatest value. 

\paragraph{Clinical settings.}
The most widely benchmarked setting for depression prediction from text is the Distress Analysis Interview Corpus--Wizard of Oz (DAIC-WOZ) \citep{gratch2014} and its extension, E-DAIC \citep{ringeval2019}. DAIC-WOZ is a set of semi-structured interviews in which a human-operated virtual agent asks participants a fixed set of questions, with transcripts labeled by self-reported PHQ-8 scores. The strongest text-only models on these corpora achieve MAEs of 3.55--3.85 on the 0--24 PHQ-8 scale \citep{schmidt2025}, including approaches that fine-tune pretrained langauge models with parameter-efficient methods \citep{lau2023}. In a related design, \citet{weber2025} predicted individual Montgomery--{\AA}sberg Depression Rating Scale (MADRS) items from 126 structured clinical interview transcripts using a fine-tuned BERT regression model. Although these results are encouraging, \citet{daicvalidity2024} showed that many DAIC-WOZ models derive their accuracy from the interviewer’s scripted prompts rather than from participants’ own language. Further, the PHQ-8 omits the suicidal ideation item of the PHQ-9, a clinically critical indicator of severity. 

Other clinical sources avoid scripted prompts but share a similar limitation. \citet{alves2025} estimated PHQ-9 scores from prescriber clinical notes for 48,594 patients, but this language is written by clinicians, not patients, and exists only because a clinical encounter took place. Studies involving psychotherapy transcripts \citep{althoff2016,ewbank2020,lalk2024} have linked conversational features to symptom severity but report modest associations ($r = 0.45$ or lower). In each of these settings, the language is generated within an existing episode of care, so models trained on it generalize to settings in which severity is already being assessed and passive inference adds the least value.

\paragraph{Non-clinical settings.}
A second line of work predicts depression from language produced outside a clinical context. Early approaches used handcrafted linguistic features such as first-person pronoun usage \citep{dechoudhury2013}, topic models \citep{resnik2015}, and absolutist language \citep{almosaiwi2018} in social media and online forum posts, with more recent work substituting pretrained transformer encoders for hand-engineered features \citep{jiang2020}. Similar dictionary-based features extracted from naturalistic text messages prospectively predicted the following week's depressive symptoms \citep{stamatis2022}. Studies that pair users' posts with self-reported PHQ scores show modest performance, with correlations of $r = 0.29$--$0.39$ for Facebook and SMS text \citep{liu2023,rai2024} and an AUC of 0.63 for Instagram posts \citep{ricard2018}. Across this literature, targets and instruments are heterogeneous, and the language often contains little explicit discussion of symptoms. Bridging clinical and non-clinical settings, \citet{shin2024} predicted binary PHQ-9 depression status ($\geq 10$) from mental-health-app diary entries using GPT-3.5 on 91 users. Naturalistic help-seeking dialogue with an AI mental health tool combines the strengths of both settings: it is produced without a clinician present, yet it consists of extended first-person accounts of mood, sleep, and functioning.

\subsection{Natural language from AI mental health tools}
A growing number of people with depression now interact with AI-powered platforms that deliver psychoeducation, cognitive-behavioral techniques, and supportive counseling through multi-turn text conversations \citep{rousmaniere2026}. These interactions generate rich, fully transcribed dialogue that captures users' self-reported experiences, emotional language, and behavioral patterns---signals plausibly related to depressive symptom severity. If successfully implemented in production systems, passive symptom monitoring based on text generated in these contexts could facilitate early detection of deterioration, adaptive treatment planning, and scalable screening that is less subject to known bias due to differential dropout and questionnaire completion \citep{teferra2024, eisele2022, hamitouche2025}. However, building these applications requires models trained on enough labelled conversations to learn severity across the full clinical range, which prior text-based work has generally lacked \citep{schmidt2025, liu2023}. Commercially deployed AI mental health tools change this. Even if only a fraction of users complete a questionnaire, the volume of use yields thousands of labelled conversations, along with a much larger pool of unlabelled ones that can expand the training data through semi-supervised methods. To date, no prior work on PHQ-9 prediction has targeted naturalistic AI mental health conversation transcripts at scale. Inferring severity from unguided therapeutic dialogue rather than from language elicited by a clinician administering a structured instrument represents a harder inference problem; however, it is the primary setting in which automated estimation is of practical use, since contexts that already deliver a structured assessment have no need for it.

\subsection{Current study}
In the present study, we estimate depression severity from users' first week of conversation with Ash, an AI mental health tool, using only the conversation text. We frame this as a regression problem: given a user's conversation history, predict their total PHQ-9 score across the full 0-27 range, rather than only whether they fall above a clinical cutoff.

Three properties of real-world labels make this problem difficult. First, labels are scarce: only users who complete the PHQ-9 can be used for training, and, as described above, those who do are neither a majority nor a random subset. Second, labels are imbalanced: in our sample, 81.8\% of users who completed the PHQ-9 scored 10 or above, leaving few examples at the low end of the scale. Third, labels are noisy: self-report questionnaires are subject to careless responding, and a user's answers can contradict what they go on to say in conversation.

Our approach is built around these constraints. We adapt an openly available large language model (Qwen 3.5 27B) \citep{qwen35} so that instead of generating text, it reads a user's conversations and outputs a single estimated PHQ-9 score. Because labeled examples are scarce, and especially scarce at the low end of the scale, we expand the training data using pseudo-labelling: a common technique in which a model's own predictions on unlabeled data are treated as additional training examples \citep{xie2020, he2020}. We begin by asking a more capable general-purpose model (Claude Opus 4.6) to estimate scores for users who did not complete the PHQ-9, and keep only the cases it judges to be low severity, where real labels are most sparse. We then use our trained model to estimate scores for more users across the full range, update those estimates each time the model improves, and average the predictions of several models to make them more stable. This expands the training set from 3,111 to 5,524 users. We compare this approach with a simpler alternative that gives extra weight to under-represented scores during training rather than adding new data. Finally, to gauge how much of the model's remaining error comes from unreliable questionnaire answers rather than from the model itself, we review a sample of transcripts for cases where what users reported on the PHQ-9 clearly contradicts what they said in conversation.

% ───2. Results ──────────────────────────────────────────────────────────────
\section{Results}

\subsection{Predicting depression severity from AI mental health conversations}
We analysed 3,953 users of Ash, an AI mental health tool, who completed the PHQ-9 before their first conversation and exchanged at least 10 messages during their first week of use (Methods). Each user's input consisted of their first-week conversation history, which ranged from 10 to 300 messages (median 80). PHQ-9 scores were elevated relative to the general population, as expected for a help-seeking sample (mean 15.95, s.d. 6.37), and 81.8\% of users scored 10 or above (Table~\ref{tab:phq_dist}). Users were split into training ($n = 3{,}111$), validation ($n = 420$) and held-out test ($n = 422$) sets, stratified by severity band so that the scarce low-severity band was represented in each slice. We fine-tuned a large language model (Qwen 3.5 27B) with a regression head to predict each user's total PHQ-9 score from conversation text alone, and expanded the training set to 5,524 users with pseudo-labels for unlabelled conversations (Methods).

Table~\ref{tab:main} shows PHQ-9 prediction performance across model variants on the held-out
test set. We report mean absolute error (MAE), Pearson $r$, AUC at the clinical threshold of 10, mean AUC at thresholds 3 through 24 (in increments of 3), and Macro-MAE. Macro-MAE is a band-weighted version of MAE, which corrects for a population that is not balanced across bands, and rewards models that predict PHQ-9 correctly  throughout the entire spectrum.
Our baseline model, the regression-fine-tuned Qwen 3.5 27B model, achieves a Macro-MAE of 4.67 and Pearson $r$ of 0.64.
Adding LLM pseudo-labels reduces Macro-MAE to 4.49, while adding the iterative pseudo-labels further reduces Macro-MAE to 4.26 and increases correlation to 0.66. 
This results support the use of pseudo-labelling as a strategy to overcome data scarcity and label imbalance. 

As a supporting test of the role of data imbalance, we additionally evaluated a Qwen 3.5 27B model trained with a reweighted loss function. 
From Table~\ref{tab:main}, we see that training with a reweighted loss reduces Macro-MAE to 4.31, which outperforms our LLM pseudo-labelling approach but not our combined LLM and iterative pseudo-labelling. 
Training with both LLM pseudo-labels and the reweighted loss has no further effect. 
These results suggest that reweighting can be an effective strategy for overcoming label imbalance, but pseudo-labelling's advantage is increasing available data scale more generally. 

Finally, having two different approaches, we also evaluate their ensemble, by averaging the predictions. We find that it produces a very modest further improvement in the performance, bringing down Macro-MAE to 4.22, and Pearson $r$ up to 0.67.

\begin{table}[h]
  \centering
  \caption{PHQ-9 prediction performance on the held-out test set ($N = 422$). AUC @3-24 refers to the mean AUC value at thresholds 3 through 24 in increments of 3. M-MAE is Macro-MAE. The two methods in the ensemble are the third and fifth rows.}
  \label{tab:main}
  \small
  \begin{tabular}{lccccc}
    \toprule
    \textbf{Model Variant} & \textbf{MAE} & \textbf{M-MAE} & \textbf{Pearson $r$} & \textbf{AUC@10} & \textbf{AUC @3-24} \\
    \midrule
    Qwen 3.5 Regression (MSE)              & 4.07 & 4.67 & 0.64 & 0.87 & 0.83 \\
    \quad + LLM pseudo-labels             & 4.26 & 4.49 & 0.63 & 0.87 & 0.82 \\
    \quad + Iterative pseudo-labels        & \textbf{3.96} & 4.26 & 0.66 & \textbf{0.89} & \textbf{0.84} \\
    \midrule
    Qwen 3.5 Regression (Weighted MAE)     & 4.26 & 4.31 & 0.65 & 0.86 & 0.83 \\
    \quad + Iterative pseudo-labels        & 4.15 & 4.30 & 0.65 & 0.87 & 0.84 \\
    \midrule
    Ensemble of both methods               & 3.98 & \textbf{4.22} & \textbf{0.67} & 0.88 & \textbf{0.84} \\
    \bottomrule
  \end{tabular}
\end{table}

\subsection{Contextual Comparison with Related Work}

Table~\ref{tab:comparison} places our results alongside those from related settings. Because the datasets, populations, clinical instruments, and task definitions differ across studies, the numbers are not directly comparable; we include them to situate our work within the broader literature and discuss the differences in Section~\ref{sec:discussion_comparison}.

\begin{table}[h]
  \centering
  \caption{Comparison with related work, grouped by outcome measure. Panel (a) compares models
    predicting PHQ-8 scores, panel (b) models predicting PHQ-9 scores, and panel (c) related work
    using other clinical measures. Settings and evaluation tasks differ across rows; see text for
    discussion.}
  \label{tab:comparison}
  \footnotesize
  \begin{tabular}{llcccc}
    \toprule
    \textbf{Study} & \textbf{Setting} & \textbf{$N$} & \textbf{MAE} & \textbf{AUC} & \textbf{$r$} \\
    \midrule
    \multicolumn{6}{l}{\textit{(a) PHQ-8}} \\
    \citet{schmidt2025} & Clinical interview (DAIC-WOZ) & 189 & 3.55 & --- & --- \\
    \citet{schmidt2025} & Clinical interview (E-DAIC)   & 275 & 3.85 & --- & --- \\
    \citet{ricard2018}  & Instagram posts               & 749 & ---  & 0.63\textsuperscript{$\dagger$} & --- \\
    \textbf{Ours} & \textbf{AI mental health conversations} & $\mathbf{3{,}953}$ & \textbf{3.76} & \textbf{0.85} & \textbf{0.64} \\
    \midrule
    \multicolumn{6}{l}{\textit{(b) PHQ-9}} \\
    % \citet{shin2024} & App diary text (binary) & 91 & --- & --- & --- \\
    \citet{alves2025} & Clinical notes        & 48,594 & n/a & 0.81 & 0.61 \\
    \citet{rai2024}   & Facebook posts        & 868    & --- & ---  & 0.39\textsuperscript{$\ddagger$} \\
    \citet{liu2023}   & Facebook / SMS text   & 120    & --- & ---  & 0.36 / 0.29 \\
    \textbf{Ours} & \textbf{AI mental health conversations} & $\mathbf{3{,}953}$ & \textbf{3.98} & \textbf{0.88} & \textbf{0.67} \\
    \midrule
    \multicolumn{6}{l}{\textit{(c) Other measures}} \\
    \citet{weber2025} & Clinical interview, MADRS (per-item) & 126 & 0.7--1.0\textsuperscript{$\S$} & --- & --- \\
    \citet{lalk2024}  & Psychotherapy transcripts            & 124 & ---  & --- & 0.45 \\
    \bottomrule
    \multicolumn{6}{l}{\footnotesize $\dagger$ User-generated text only, PHQ-8 $\geq$ 15; 0.73 when followers' comments are added.} \\
    \multicolumn{6}{l}{\footnotesize $\ddagger$ Model trained and tested on White participants; $r = 0.13$ on the Black test set.} \\
    \multicolumn{6}{l}{\footnotesize $\S$ Per-item MAE on the MADRS 0--6 scale; not directly comparable to PHQ totals.}
  \end{tabular}
\end{table}

\subsection{Binary Screening and Multi-Threshold Discrimination}

Table~\ref{tab:binary} shows depression screening performance at the PHQ-9~$\geq 10$ and $\geq 15$ thresholds. Note that at the traditional clinical threshold of 10, the sensitivity-specificity trade-off is heavily influenced by the occurrence rates in our ecological sample, which simply did not contain many users with scores below 10. Because of that same skew, even an uninformative always-positive classifier attains high precision and F1 at the PHQ-9 $\geq$ 10 threshold: its precision equals the prevalence $P$, and its F1 equals $2P/(1+P)$ (0.92 at $\geq 10$, 0.72 at $\geq 15$). That always-positive classifier obviously gets progressively worse at higher thresholds, unlike our model.

\begin{table}[h]
    \centering
    \caption{Binary depression screening performance vs.\ an always-positive
      baseline (predicts every user as a case). Given the high base rates in this
      help-seeking cohort, the always-positive classifier already attains high
      PPV and F1, so the model's gains are best read in specificity and AUC.}
    \label{tab:binary}
    \small  
    \begin{tabular}{llcccccc}
      \toprule
      \textbf{Threshold} & \textbf{Method} & \textbf{AUC} & \textbf{Sens.} & \textbf{Spec.} & \textbf{PPV} & \textbf{NPV} & \textbf{F1} \\
      \midrule
      \multirow{2}{*}{\textbf{10}} 
        & Model           & 0.88 & 0.92 & 0.64 & 0.91 & 0.67 & 0.92 \\
        & Always-positive & 0.50 & 1.00 & 0.00 & 0.80 & ---   & 0.89 \\
      \midrule
      \multirow{2}{*}{\textbf{15}} 
        & Model           & 0.84 & 0.72 & 0.81 & 0.84 & 0.68 & 0.78 \\
        & Always-positive & 0.50 & 1.00 & 0.00 & 0.59 & ---   & 0.74 \\
      \bottomrule
    \end{tabular}
  \end{table}

\begin{figure}[h]
    \centering
    \includegraphics[width=0.82\linewidth]{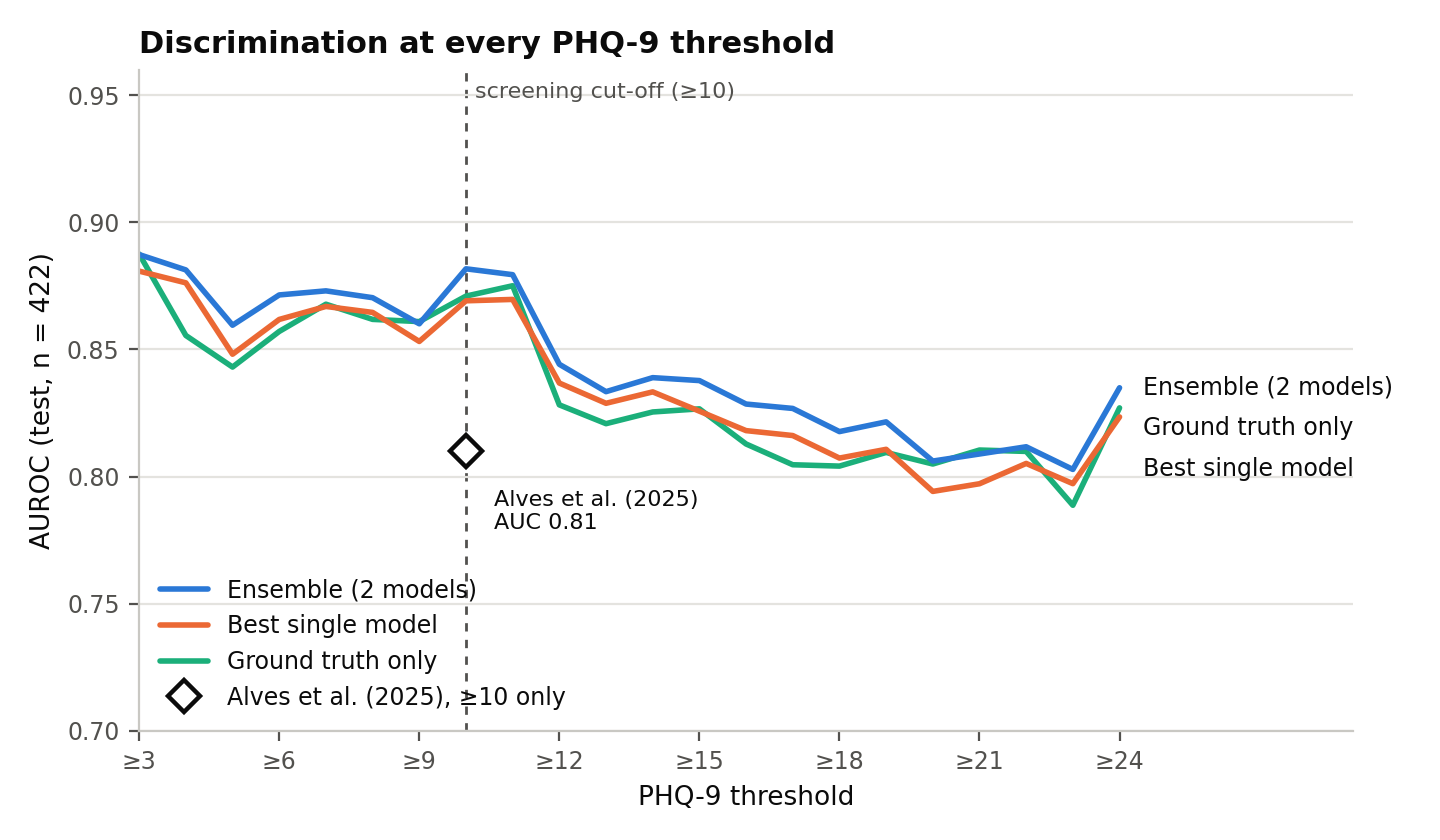}
    \caption{AUC for binary classification at each PHQ-9 threshold ($\geq k$),
      across model iterations. The final ensemble maintains high AUC
      (0.80-0.89) across the full clinical range. The diamond is an
      external reference point---Alves et al.\ (2025), AUC~$=0.81$ for binary
      PHQ-9~$\geq 10$ from prescriber clinical notes; settings and populations
      differ, so it is a contextual baseline only.}
    \label{fig:auc_thresholds}
\end{figure}

A key advantage of our approach is that strong discrimination extends across the full PHQ-9 severity range. Fig~\ref{fig:auc_thresholds} shows AUC at every threshold from $\geq 3$ to $\geq 24$; AUC remains above 0.80 at every point. This contrasts with binary screening models optimised and evaluated at a single cutoff. The model can therefore support clinical decisions at multiple severity levels---identifying users who cross from mild into moderate (PHQ-9~$\geq 10$) or from moderate into moderately severe (PHQ-9~$\geq 15$), each of which
may warrant a different clinical response.

\subsection{Error Analysis}
\label{sec:error-analysis}

Table~\ref{tab:confusion} shows a 5$\times$5 confusion matrix with rows representing true severity band and columns representing predicted severity band (predicted score binned into the same five standard ranges). Per-band accuracy and MAE are shown alongside.

\begin{table}[h]
  \centering
  \caption{5$\times$5 confusion matrix by severity band (ensemble, $N = 422$). Rows = true band; columns = predicted band (predicted total score binned into standard ranges). Per-band in-band accuracy and MAE are shown in the rightmost columns.}
  \label{tab:confusion}
  \small
  \begin{tabular}{lrrrrrrrr}
    \toprule
    \textbf{True $\backslash$ Pred} & \textbf{Min} & \textbf{Mild} & \textbf{Mod} & \textbf{MoSev} & \textbf{Sev}
      & N & \textbf{recall} & \textbf{MAE} \\
    \midrule
    Minimal             &  \textbf{1} & 18 & 9 &  1 &  0 & 29 & 0.03 & 6.43 \\
    Mild               &  0 & \textbf{31} & 18 & 5 &  0 & 54 & 0.57 & 3.13 \\
    Moderate           &  1 & 13 & \textbf{46} & 23 & 9 & 92 & 0.50 & 3.37 \\
    Mod.\ severe       &  0 &  7 &  35 & \textbf{54} & 24 & 120 & 0.45 & 3.12 \\
    Severe             &  0 &  2 &  18 &  56 & \textbf{51} & 127 & 0.40 & 5.04 \\
    \bottomrule
  \end{tabular}
\end{table}

Prediction recall has an inverse U-shaped curve: the model is best at the mild and moderate bands, and worse at the minimal and severe ones. Most errors are adjacent-band misclassifications (e.g., minimal predicted as mild, moderate predicted as moderately severe), with few large cross-band errors: the within-1-band accuracy is 88\%. The low recall for minimal cases reflects the fact that the model, trained predominantly on a high-severity cohort (81\% cases), tends to predict scores in the moderate range even for users with few symptoms. Fig. 2 shows this pattern across the full score range: predictions are pulled toward the sample mean, overestimating low scores and underestimating high ones, consistent with the reduced recall in both the minimal and severe bands (Table 4).

\begin{figure}[h]
    \centering
    \includegraphics[width=0.7\linewidth]{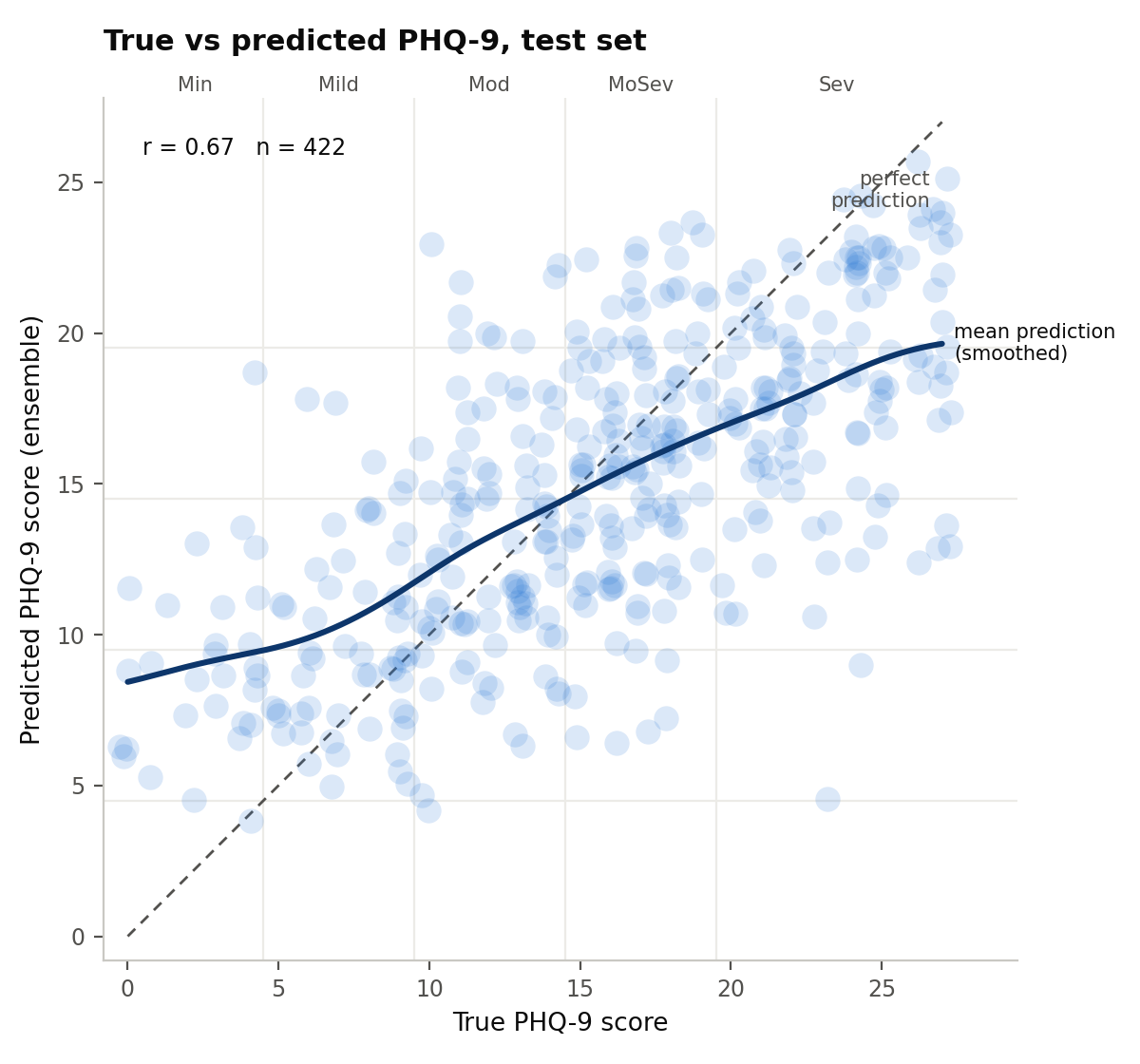}
    \caption{Predicted vs. true scores, showing the classic reversion to the mean pattern.}
    \label{fig:calibration}
\end{figure}

\subsection{Label Noise Analysis}

To provide additional context on the degree of noise in the PHQ-9 scores, Table~\ref{tab:transcript-audit} shows the results of our LLM-assisted transcript audit. 
According to the frontier LLM judge, there is direct evidence in the transcript that contradicts the overall score in 10.1\% of transcripts, rising to 33.3\% for users in the minimal band. 
7.1\% of users respond with ``straight line'' responses, where a user responds to every questionnaire item identically, and 1.7\% of users respond to every item with a zero. 
While these all-zero and straight line responses do not conclusively identify label noise, they do indicate possible satisficing. 
Rates of transcript contradiction and straight-line responses are highest in minimal and severe bands and lowest in the moderate bands, mirroring the distribution of prediction error in Section~\ref{sec:error-analysis}.
We interpret these results as suggesting that predicting PHQ-9 scores from naturalistic data is further complicated by the presence of significant label noise in the scores themselves. 

% ─── 6. Discussion ───────────────────────────────────────────────────────────
\section{Discussion}

\subsection{Summary of Findings}
We demonstrate that PHQ-9 depression severity can be predicted from naturalistic AI mental health conversations with macro-MAE~4.22, Pearson $r = 0.67$, and AUC~0.88 at the clinical screening threshold of PHQ-9~$\geq 10$. Critically, creating pseudo-labels to expand and balance the training data helped increase the accuracy of the final model. The predictions discriminate depression severity across the full PHQ-9 range, with AUC above 0.80 at every threshold from $\geq 3$ to $\geq 24$, enabling graded clinical distinctions rather than binary case detection.

\subsection{Contextual Comparison with Related Work}
\label{sec:discussion_comparison}

Because no prior work uses the same setting, instrument, and evaluation protocol, the numbers in Table~\ref{tab:comparison} are contextual.

Compared to \citet{shin2024}, who predict PHQ-9 (binary; above/below cutoff) from diary entries in a mental health app, our sensitivity (0.92 vs.\ 0.62) is substantially higher at the cost of lower specificity (0.64 vs.\ 0.95). For depression screening, missing a case (false negative) is generally considered more harmful than a false alarm, so this trade-off may be clinically appropriate. Our lower specificity partly reflects calibration toward the help-seeking cohort (81.8\% of users); raising the decision threshold above 10 yields more balanced sensitivity/specificity without retraining. Similar to \citet{weber2025}, we fine-tune LLMs on clinical text and attach a regression head over the backbone's hidden state. The main difference is the source of the language: their MADRS items are elicited by a clinician through structured prompts, whereas our PHQ-9 score must be inferred from unguided therapeutic dialogue. Our setting is not only a harder inference problem but arguably the one that matters in practice---a model that depends on clinician-administered prompts cannot operate where no clinician is present, which is precisely where passive, scalable screening is needed.

\subsection{Cross-Severity Granularity}
Standard binary screening tools are calibrated at a single severity threshold and cannot
distinguish between, for example, a user with PHQ-9~11 (moderate) and PHQ-9~18 (moderately
severe). Our model's AUC exceeds 0.80 at every threshold across the clinical spectrum, meaning it can support graded clinical decisions: triaging users into severity bands, tracking within-user changes over time, or triggering escalation when a user crosses from one severity category into the next. This granularity is a direct consequence of predicting the full continuous score rather than optimising for a single binary cutoff.

\subsection{Clinical Implications}
If conversation-based estimates prove sufficiently accurate in prospective use, they could support several applications, each of which depends on properties our evaluation addresses only in part.

\paragraph{Passive symptom monitoring.}
Individuals could receive ongoing symptom severity estimates without completing questionnaires, reducing assessment burden and increasing measurement frequency. For example, measurement-based care---routine tracking of symptoms with validated measures---has been shown to improve outcomes \citep{fortney2017, kumar2024}, but is rarely implemented in practice due to the logistical strain and participant burden of repeated questionnaires \citep{lewis2019}. This problem is compounded by low completion rates and attrition of the populations most at risk \citep{pratap2020, hamitouche2025}. Estimates inferred from conversations that people are already having avoid both problems, and would be available for more users, not just those who complete the surveys. Our results support this use at the level of severity bands. 88\% of predictions fall within one band of the true score, and AUC exceeds 0.80 at every threshold.

\paragraph{Early deterioration detection.}
Sudden increases in predicted depression severity are key moments for adjusting treatment or escalations to higher levels of care. In psychotherapy, routine outcome monitoring (ROM) systems that flag patients who are not progressing as expected can reduce deterioration and dropout \citep{lambert2018, dejong2021}. Conversation-based estimates could serve a similar function by flagging users whose predicted severity rises for human review or prompting a recommendation of professional care. Two features of our results limit this use at present. The model compresses predictions toward the middle of the scale, so recall in the severe band is 0.40. And a 5-point change on the PHQ-9 is generally considered clinically meaningful \citep{lowe2004}, which is close to the model's mean absolute error. Before such alerts are deployed, the model will need validation against within-user change and better calibration at the upper end of the scale.

\paragraph{Treatment personalization.}
Adaptive interventions adjust the type or intensity of support according to a person's changing state \citep{nahumshani2018}. Within an AI mental health tool, severity estimates could inform such decisions, for example by shifting strategy when predicted scores suggest a user is not improving, or by recommending a higher level of care.

\paragraph{Population-level screening.}
Depression remains under-detected in routine care \citep{garcia2022, fekadu2022} despite screening being recommended for the general adult population \citep{uspstf2023}. Aggregated estimates could identify groups at elevated risk without adding measurement burden, and aggregation reduces the effect of individual-level error. However, Ash users, and the subset who complete the PHQ-9, are self-selected in ways that are difficult to measure. Language-based depression models can also perform unevenly across demographic groups \citep{rai2024}. Population use will therefore require validation on other platforms and populations, and reporting of performance across subgroups.

\subsection{Conclusion}
We present, to our knowledge, the first large-scale study of PHQ-9 score prediction from AI-mediated therapy conversations. By fine-tuning Qwen 3.5 27B with a regression head and augmenting labeled data through pseudo-labelling (521 conversations from a frontier reasoning LLM and 1,892 from an iteratively trained model), we achieve macro-MAE~4.22, Pearson $r = 0.67$, and AUC~0.88 on a held-out test set of 422 users, with AUC above 0.80 at every PHQ-9 severity threshold from $\geq 3$ to $\geq 24$. Our results suggest that conversational text generated in AI mental health sessions contains sufficient signal to estimate depressive symptom severity, opening the door to passive monitoring, adaptive treatment, and scalable mental health screening without requiring users to complete questionnaires at all. Future work in this direction could evaluate generalization across platforms, predict other clinical constructs, and explore temporal dynamics of predicted severity over the course of treatment.

% ─── 6. Methods ──────────────────────────────────────────────────────────────
%TC:ignore
\section{Methods}

\subsection{Ethics Approval}
The study was reviewed by the Biomedical Research Alliance of New York (BRANY) independent Institutional Review Board (IRB) and determined to be exempt under category 4(ii) (BRANY IRB File \# 26-081-2393).

\subsection{Study Population}
Our dataset is drawn from Ash, an AI-powered mental health platform developed by Slingshot AI. Users engage in multi-turn text-based conversations with an LLM-based mental health tool trained to draw from evidence-based strategies such as cognitive behavioral therapy (CBT) and acceptance and commitment therapy (ACT).
We collected 4,976 PHQ-9 scores prior to users' first conversation with Ash.
We include users who completed the full PHQ-9 and had at least one conversation with Ash during the first seven days of use, with a minimum of 10 messages.
This leaves us with a final sample of 3,953 users, which we split using stratified sampling into train, validation, and test partitions of 3,111, 420, and 422 users respectively.

\subsection{Data}
\paragraph {PHQ-9 Assessments}

The Patient Health Questionnaire-9 \citep[PHQ-9;][]{kroenke2001} is among the most widely used self-report instruments for screening and monitoring depressive symptom severity.
Each of the nine items maps to a DSM-5 criterion for major depressive disorder and is scored from 0 (``not at all'') to 3 (``nearly every day''), yielding a total score between 0 and 27.
Standard clinical severity thresholds classify scores of 0--4 as minimal, 5--9 as mild, 10--14 as moderate, 15--19 as moderately severe, and 20--27 as severe depression \citep{kroenke2001}.

PHQ-9 scores in our sample are substantially elevated relative to the general population, with mean~15.95 (SD~6.37) and median~16, as expected for a help-seeking cohort.
Table~\ref{tab:phq_dist} shows the distribution using the standard clinical severity bands.
The most frequently endorsed items are guilt/worthlessness (item~6, mean 2.14), fatigue (item~4, mean 2.13), and depressed mood (item~2, mean 2.01).
Psychomotor disturbance (item~8, mean 1.22) and suicidal ideation (item~9, mean 1.13) are the least endorsed on average, though suicidal ideation is still endorsed at a non-trivial rate: 60.6\% of responses are above 0.

\begin{table}[h]
  \centering
  \caption{PHQ-9 severity distribution using standard clinical severity bands
    \citep{kroenke2001}.}
  \label{tab:phq_dist}
  \small
  \begin{tabular}{lcrr}
    \toprule
    \textbf{Severity} & \textbf{Range} & \textbf{$N$} & \textbf{\%} \\
    \midrule
    Minimal           & 0--4   &  125 &  3.2 \\
    Mild              & 5--9   &  596 & 15.1 \\
    Moderate          & 10--14 & 912 & 23.1 \\
    Moderately severe & 15--19 & 1,066 & 27.0 \\
    Severe            & 20--27 & 1,254 & 31.7 \\
    \midrule
    \textbf{Above clinical threshold} ($\geq$10) &   & \textbf{3,232} & \textbf{81.8} \\
    \bottomrule
  \end{tabular}
\end{table}

\paragraph{Conversation Data}
\label{sec:data_conversations}

Each user's input is constructed from their Ash conversation messages during the first seven days of use.
Messages are formatted as alternating user and assistant turns. Users with fewer than 10 messages were excluded.
When a user's message history exceeded 300 messages, we subsampled to exactly 300 by drawing at evenly-spaced indices across the full first-week window, preserving the temporal arc of the conversation rather than truncating.
In the final dataset, conversation length ranges from 10 to 300 messages (mean~119.1, median~80).

\subsection{Data Analysis Methods}
\subsubsection{PHQ-9 Regression Model}
\label{sec:finetuning}

We remove the token prediction head and attach a regression head to a Qwen 3.5 27B backbone.
The regression head takes the hidden state of the final transformer layer and maps it through a linear layer to a single continuous PHQ-9 total score prediction.

Each user's input consists of their first-week conversation history, formatted as alternating \texttt{user} and \texttt{assistant} turns with the model's chat template, and tokenized with a maximum sequence length of 16,384 tokens.
We use low rank adaptation (LoRA) to improve the efficiency of fine-tuning \citet{hu2022lora}, optimising a mean squared error (MSE) loss on the total-score regression target for 5 epochs using an effective batch size and early stopping according to macro-weighted MAE (see Section~{\ref{sec:evaluation_metrics}}) on the validation set. 
Parameters are optimised using AdamW with a cosine learning rate schedule with initial learning rate $5\times10^{-5}$ and warm-up fraction of 0.1.
See Appendix \ref{app:training} for full training details. 

\subsubsection{Pseudo-Label Generation}
\label{sec:pseudo-labels}

To supplement the 3,111 labeled users in our training set, and to correct the label imbalance that stems from the help-seeking nature of the population, we generate pseudo-labels for an additional 2,413 unlabeled Ash conversations, bringing the total training set to 5,524 samples.
Pseudo-labels are generated from two sources: a frontier reasoning LLM  with prompted in a few-shot setting, and an intermediate fine-tuned regression model.

\paragraph{Frontier LLM pseudo-labels.}
To obtain extra training data, we instructed Claude Opus 4.6 to estimate the PHQ-9 score for unlabelled conversation transcript. Sampling 3,000 random users, filtering out the ones that had exchanged less than 10 messages with Ash, and selecting the ones that were estimated to have PHQ-9 scores below 10, we obtained pseudo-labels for 521 extra users. The choice to only retain low-severity pseudo-labels here was made to balance out their significant underrepresentation in our ground truth dataset. Details are presented in Appendix~\ref{app:opus-pseudo-labelling}.

\paragraph{Iterative model pseudo-labels.}
After an initial round of supervised fine-tuning on the ground-truth and LLM-labeled data (Section~\ref{sec:finetuning}), we find a lower MAE, but also slightly lower Pearson $r$ and AUC scores. We use the trained model itself to generate pseudo-labels for an additional 1,892 unlabeled users, sampled and filtered in the same way as for the Opus round. During this round of training we make use of iteratively pseudo-labelled users across all severity bands (i.e. not just the low-severity ones) yielding a combined pseudo-labeled set of 2,413 users and a total training set of 5,524. In Table \ref{tab:main} we refer to this set as "iterative pseudo-labels".

Training a new model on the iterative pseudo-labels improves performance. Re-labelling the same set of pseudo-labelled users with that improved model yields a modest further improvement (iterative pseudo-labels). Finally, an ensemble of two models trained with different methods on the same dataset, simply averaging their predicted scores, gives the best test set performance we have seen.

\subsubsection{Loss Reweighting}

We additional test a reweighted loss function an alternative strategy for countering class imbalance, without correcting for general data scarcity.
Here, we adjust a mean-absolute error loss function by scaling the weight of each sample's loss by the inverse of the number of samples in its true band, such that our generally scarce minimal-band users would produce a higher loss than our more severe users.
In Table \ref{tab:main} we refer to this set as "Weighted MAE", we test it with and without additional frontier LLM pseudo-labels.  
This is exactly equivalent to training on the Macro-MAE metric we describe in Section~\ref{sec:evaluation_metrics}.

\subsection{Evaluation Metrics}
\label{sec:evaluation_metrics}

We evaluate our model using the following metrics on the held-out test set:
\begin{itemize}[leftmargin=*]
  \item \textbf{Mean Absolute Error (MAE):} Average absolute difference between predicted and true PHQ-9 scores.
  \item \textbf{Macro-MAE:} The weighted average of the per-band MAE, where each band is weighted inversely proportionally to its number of samples. This rewards models that are uniformly good across all severity bands. 
  \item \textbf{Pearson correlation ($r$):} Linear correlation between predicted and true scores.
  \item \textbf{AUROC:} Treating our regression model as a binary classifier for clinical/sub-clinical users ($\geq$ 10), we report the area under the receiver operating characteristic curve (AUROC, hereafter AUC), sensitivity, specificity, and F1 at the standard PHQ-9~$\geq 10$ clinical threshold. We also sweep all thresholds from $\geq 3$ to $\geq 24$ to assess discrimination across the full severity range.
\end{itemize}

\subsubsection{Label Noise Analysis}
\label{sec:methods-label-noise}

We hypothesize that a contributing factor to the difficulty of PHQ-9 prediction is noise in the scores themselves. 
To evaluate this, we conducted an LLM-assisted audit of a random sample of transcripts to check indicators of noise.
We sampled N=750 transcripts from the training partition of our data, and used a frontier LLM (Claude Opus 4.8) with a custom prompt to look for direct passages where the transcript contradicted the user's PHQ-9 score.
See Appendix~\ref{app:label-noise-analysis} for full details. 
For example, this audit would reveal cases where the user responded with a zero (``not at all'') to the PHQ item about sleep problems, but then directly mentioned sleep-related symptoms in their conversation. 
We report contradiction rates from the LLM audit, alongside the proportion of users who responded with a zero to all PHQ-9 items, or who responded to all items with an identical score (i.e., responding in a ``straight line''), both of which are evidence of possible satisficing. 
% ─── Declarations ──────────────────────────────────────────────────────────────
\section*{Data Availability}
The conversation transcripts and PHQ-9 responses analysed in this study were collected from users of Ash, a commercial AI mental health application, and contain sensitive personal health information. To protect user privacy, and in accordance with the terms under which the data were collected and the IRB exemption determination (BRANY IRB File \#26-081-2393), these data cannot be made publicly available. Aggregate, non-identifiable summary statistics underlying the tables and figures are available from the corresponding author upon reasonable request; requests will be reviewed by Slingshot AI to ensure compliance with privacy obligations.
% Delete the last sentence if the team won't honor such requests

\section*{Code Availability}
The code used for model training and evaluation is proprietary and not publicly available. The prompts used for pseudo-labelling and label-noise auditing are provided in full in Appendices~A and~C, and all training hyperparameters are reported in Appendix~B.
% Not on the editor's list but generally required by Nature Portfolio journals.
% If code will be released, replace with: The training and evaluation code is available at \url{[repository URL]}.

\section*{Funding}
This work was funded by Slingshot AI. No external funding was received.
% Add any relevant author grant support

\section*{Author Contributions}
O.T., S.J.B., Z.Z.\ and T.S.\ developed the modelling approach, conducted model training, and performed the analyses. O.T.\ and S.J.B.\ designed the pseudo-labelling and loss-reweighting experiments. E.K.\ contributed to the clinical framing, the review of related literature, and manuscript revision. T.D.H.\ and C.A.S.\ conceptualized the study and supervised the project. C.A.S.\ oversaw the research ethics review and data access. All authors contributed to writing and revising the manuscript and approved the final version.
% Best guess from author order -- confirm roles with each author

\section*{Competing Interests}
All authors are employees of Slingshot AI, the developer of Ash, the AI mental health application from which the study data were drawn, and hold equity in the company. Slingshot AI has a commercial interest in the development of passive symptom-monitoring capabilities such as those described in this study. The authors declare no other competing interests.
% Confirm the equity clause applies to every author

% ─── References ──────────────────────────────────────────────────────────────
\bibliographystyle{plainnat}

\bibliographystyle{plainnat}

% ─── Appendix ──────────────────────────────────────────────────────────────
\appendix
\section*{Appendix}
\section{Frontier LLM Pseudo-labelling}\label{app:opus-pseudo-labelling}
To produce the frontier LLM pseudo labels, we prompted a Claude Opus 4.6 model to predict overall PHQ-9 scores.
Given a user's transcript, we used the following prompt, with default inference parameters:
\begin{lstlisting}
You are an expert clinical psychologist specialising in depression assessment using the Patient Health Questionnaire-9 (PHQ-9), a validated 9-item scale scored 0–27.

Severity bands: 
  0–4   Minimal depression
  5–9   Mild depression
  10–14 Moderate depression
  15–19 Moderately severe depression
  20–27 Severe depression

Your task: read a conversation between a user and an AI therapist, then
estimate the user's current PHQ-9 score. Consider: \\
  • Symptom language (hopelessness, worthlessness, sleep, appetite, energy) \\
  • Emotional tone and affect intensity \\
  • Functional impairment described \\
  • Clinical vocabulary and self-awareness \\
  • Persistence and pervasiveness of depressive themes

Calibration matters — avoid compressing scores toward the middle.

Below are calibration examples with clinician-verified PHQ-9 scores.

[FEW-SHOT EXAMPLES]

For each new conversation provided, respond with ONLY valid JSON:
  {"phq9\_score": <number 0-27>, "reasoning": "<1-2 sentences>"}

Do not include any other text outside the JSON object.
\end{lstlisting}

\section{Training Details}\label{app:training}

\begin{table}[h]
  \centering
  \caption{Settings shared by every run.}
  \label{tab:hyperparams}
  \small
  \begin{tabular}{ll}
    \toprule
    \textbf{Setting} & \textbf{Value} \\
    \midrule
    Base model                 & Qwen3.5-27B \\
    Quantisation               & 4-bit NF4, double quantisation, bfloat16 compute \\
    Adapter                    & LoRA, rank 128, $\alpha = 128$, dropout 0.1 \\
    Adapted modules            & q, k, v, o, gate, up, down projections; regression head trained in full \\
    Output                     & Single linear head, raw PHQ-9 total (0--27) \\
    Maximum input length       & 16,384 tokens (right-truncated) \\
    Optimiser                  & AdamW (8-bit), weight decay 0.01 \\
    Learning rate              & $5 \times 10^{-5}$, cosine decay, 10\% linear warm-up \\
    Batch size                 & 1 per step $\times$ 8 accumulation steps (effective 8) \\
    Epochs                     & 3.3-4.6 (different runs had different best checkpoints) \\
    Hardware                   & 1 $\times$ NVIDIA H100 80\,GB per run \\
    \bottomrule
  \end{tabular}
\end{table}

All models in Table~\ref{tab:main} share the same architecture, as well as most optimisation settings. The shared settings are listed in Table~\ref{tab:hyperparams}. Runs only differ in their training data: which pseudo-labeled set, if any, and in how the checkpoint was selected: by MSE or macro-MAE on the validation set.

\paragraph{Model.}
We fine-tune Qwen3.5-27B as a sequence regressor: a single linear output (the \texttt{score} head) on the final token's
hidden state, predicting the PHQ-9 total score on its raw 0--27 scale. The base model is loaded in 4-bit NF4 with
double quantisation and bfloat16 compute (QLoRA), and we train LoRA adapters on every attention and MLP projection
(\texttt{q,k,v,o,gate,up,down}) together with the full regression head. Gradient checkpointing keeps a 16,384-token
input within a single 80\,GB H100.

\paragraph{Input.}
Each example is one user's conversation window, rendered with the model's chat template. Inputs longer than 16,384
tokens are truncated on the right, as at inference time.

\paragraph{Held-out data.}
All runs share the same 842-user held-out set. We split it at the user level into a 420-user selection half and a 422-user test half; checkpoints are chosen on the selection half and every number in
Table~\ref{tab:main} is computed on the test half.

\paragraph{Losses.}
The MSE runs minimise squared error on the raw score. The weighted-MAE runs minimise absolute error with each training
row weighted by the inverse frequency of its true severity band (minimal 0--4, mild 5--9, moderate 10--14, moderately
severe 15--19, severe 20--27), $w_k \propto n_k^{-1}$, normalised to a mean weight of one so that the learning rate keeps
its meaning. With these weights every band contributes equally to the loss, which makes the training objective the
macro-averaged MAE we report. The weighting is strong on ground truth alone, where minimal-band users are scarce (a 14.9-fold
spread between the largest and smallest band weight, 39\% effective sample size), and mild once pseudo-labels are
added, since the pseudo-labelled pool is rich in low scores (2.5-fold, 78\%).

\paragraph{Checkpoint selection.}
We evaluate on the validation set every 200 optimiser steps and save every prediction. For each run we take the
checkpoint with the lowest macro-MAE on the 420-user selection half; predictions are clipped to 0--27 but not rounded. The selected checkpoint and corresponding epoch for each run are listed in Table 7.

\begin{table}[h]
  \centering
  \caption{Selected checkpoint for each run (lowest macro-MAE on the validation half).}
  \label{tab:checkpoints}
  \small
  \begin{tabular}{lrr}
    \toprule
    \textbf{Model variant} & \textbf{Checkpoint (step)} & \textbf{Epoch} \\
    \midrule
    Regression (MSE)                 & 1,800 & 4.6 \\
    \quad + LLM pseudo-labels        & 1,600 & 3.4 \\
    \quad + Iterative pseudo-labels  & 2,400 & 3.3 \\
    \midrule
    Regression (weighted MAE)        & 1,600 & 4.1 \\
    \quad + Iterative pseudo-labels  & 2,600 & 3.5 \\
    \bottomrule
  \end{tabular}
\end{table}

\paragraph{Compute.}
Each run used one NVIDIA H100 (80\,GB). Training took 14\,h on ground truth alone, 17\,h with LLM pseudo-labels and
32\,h with iterative pseudo-labels.

\section{Label Noise Analysis}\label{app:label-noise-analysis}

For the label noise analysis described in Section~\ref{sec:methods-label-noise}, we used a Claude Opus 4.8 model to grade transcripts according to whether they contradict the user's reported PHQ item responses.
The LLM was given a single user's transcript, the user's corresponding item-level responses and overall severity band, and the following prompt, with default inference parameters:

\begin{lstlisting}
You are a clinical assessor with expertise in depression screening. You audit whether a user's self-reported PHQ-9 score is consistent with what they show in conversation. You are deliberately two-sided: you must look for evidence against the score as hard as you look for evidence in favour of it.

Below is a window of conversation between a user and an AI therapist, taken from around the user's intake. Message timestamps are included.

--- Session started {YYYY-MM-DD HH:MM} UTC ---
[{HH:MM}] user: {content}
[{HH:MM}] therapist: {content}
...

At intake, this user submitted a PHQ-9 with:
- total: {total} / 27 (severity: {severity})
Per-item scores (0 = not at all ... 3 = nearly every day, over the last two weeks):
- {item label}: {score}
...

Audit whether the transcript is consistent with this score:
- supporting_evidence: the strongest evidence in the transcript that supports a score in this severity range (shortest quote or cue description; "none found" if there is genuinely none).
- contradicting_evidence: the strongest evidence that points to a clearly different severity (same rules; "none found" if none).
- verdict: `supported' if the transcript evidence is consistent with this severity range; `contradicted' if it points to a clearly different severity; `unverifiable' if the transcript lacks the evidence to say either way.
- rationale: 2-3 sentences weighing the two sides.
\end{lstlisting}

\begin{table}[H]
  \centering
  \caption{Share of PHQ-9 labels that the conversation transcript contradicts, by severity band, for 750 training users. Straight-lined: all nine items scored the same. All zeroes: every item scored 0, a subset of straight-lined.}
  \label{tab:transcript-audit}
  \begin{tabular}{lrrrr}
    \toprule
    \textbf{Severity band} & \textbf{$n$} & \textbf{Contradicted} & \textbf{Straight-lined} & \textbf{All zeroes} \\
    \midrule
    Minimal           &  51 & 33.3\% & 25.5\% & 25.5\% \\
    Mild              & 180 & 20.6\% &  6.1\% &  0.0\% \\
    Moderate          & 178 &  2.8\% &  0.0\% &  0.0\% \\
    Moderately severe & 167 &  2.4\% &  3.6\% &  0.0\% \\
    Severe            & 174 &  7.5\% & 13.2\% &  0.0\% \\
    \midrule
    Total             & 750 & 10.1\% &  7.1\% &  1.7\% \\
    \bottomrule
  \end{tabular}
\end{table}

%TC:endignore
\end{document}